\documentclass[lettersize,journal]{IEEEtran}

\usepackage[protrusion=false,expansion=false]{microtype}
\usepackage{graphicx}
\usepackage{capt-of}
\usepackage{amsmath}
\usepackage{amssymb}
\usepackage{booktabs}
\usepackage{array}
\usepackage{siunitx}
\usepackage{url}
\usepackage[nocompress]{cite}
\usepackage{xcolor}
\usepackage[capitalise,noabbrev]{cleveref}
\crefname{figure}{Fig.}{Figs.}
\Crefname{figure}{Fig.}{Figs.}

\graphicspath{{figures/}{figures/pdf/}{figures/png/}}

\title{\textls[-10]{Autonomous Track Racing with a Commercially Available Humanoid:}\\
System Design and Real-World Deployment
}

  \author{Koki Inami$^{1}$, Takashi Sato$^{2}$, Hiroyuki Oka$^{2}$, Shota Takizawa$^{1}$, Yuya Takahashi$^{1}$,\\
  Amon Suzuki$^{2}$, Hayato Ono$^{2}$, Kohtaro Ihara$^{2}$, Yuki Noka$^{2}$, and Shohei Matsugu$^{1}$%
  \thanks{$^{1}$GMO Internet Group, Inc., Tokyo, Japan}%
  \thanks{$^{2}$GMO Various Robotics Inc., Tokyo, Japan}%
  }

\IEEEaftertitletext{\begin{minipage}{\textwidth}
  \centering
      \includegraphics[width=0.88\textwidth]{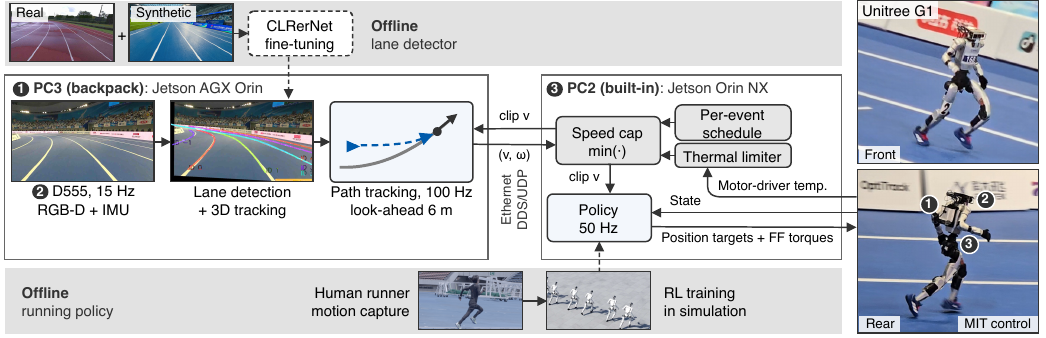}
  \vspace{-5pt}
  \captionof{figure}{Overview of the autonomous running system deployed on a commercially available Unitree G1
  for the 100 m, 400 m, and 1500 m events at the World Humanoid Robot Games 2026.
  The outlined boxes show the two onboard computers we programmed.
  PC3 turns the head-camera images into the command $(v,\omega)$,
  and PC2 executes this command with the learned running policy.
  The shaded bands are offline training, and \cref{sec:system} describes the numbered components.
  The camera images show a single instant of the 1500\,m race: the raw image and the image with the
  tracked lanes reprojected with their IDs.}
  \label{fig:system}
\end{minipage}
\vspace{0.2\baselineskip}
}

\makeatletter
\def\@IEEEpubidpullup{3\baselineskip}
\makeatother
\IEEEpubid{\parbox[b]{\columnwidth}{\footnotesize
This work has been submitted to the IEEE for possible publication.
Copyright may be transferred without notice, after which this version may no longer be accessible.}%
\hspace{\columnsep}\makebox[\columnwidth]{}}
\begin{document}
\bstctlcite{IEEEexample:BSTcontrol}
\maketitle

\begin{abstract}
Learning-based methods now let humanoids run robustly, and they have begun to compete in athletic events.
Yet few reports describe a complete autonomous humanoid running system in an actual race,
where the robot must follow a lane at a new venue, pace itself within hardware limits, and finish without intervention.
We report such a system deployed on the Unitree G1, a commercially available general-purpose humanoid,
in the 100 m, 400 m, and 1500 m events of the World Humanoid Robot Games 2026.
The system integrates vision-based lane perception, path tracking,
and a running policy trained by reinforcement learning with a human motion prior.
To address the challenges of humanoid running,
we stabilize head-camera images against body motion and use IMU-assisted leg odometry to bridge flight phases.
To manage motor heating,
we penalize hip-pitch torque during training, add heat-dissipation plates,
and apply temperature-based speed limiting with event-specific pacing.
The robot placed 13th of 74 in the 100\,m, 10th of 44 in the 400\,m first round before falling in the semifinal, and seventh of 35 in the 1500\,m.
In this deployment, thermal margin set the pace in every event, and the semifinal fall followed a lane-detection error.
\end{abstract}


\section{Introduction}
\label{sec:intro}

\IEEEPARstart{E}{ntering} a track event demands more of an autonomous humanoid than generating a running gait.
The robot must recognize and follow a lane at a venue absent from its real training images, negotiate the curves, select a pace that its hardware
can sustain, and complete the attempt without intervention or a second chance.
This paper describes one such system (\cref{fig:system}) as deployed in an
international competition.
Because the races allowed no restart and a fall could damage the robot, we could not run extensive on-robot ablations.
Instead, we focus on system integration, design decisions, race results, and observed failures.

Lane detection is well established in autonomous driving~\cite{pan2018scnn,qin2020ufld,wang2022ganet,zheng2022clrnet,honda2024clrernet}, while onboard vision has supported terrain traversal and navigation in legged robots~\cite{agarwal2023egocentric,fu2022coupling}.
Learning-based locomotion has advanced from quadrupeds to humanoid running~\cite{hwangbo2019anymal,rudin2022minutes,radosavovic2024humanoid,he2025asap,siekmann2021gaits,li2025cassie,wei2026sprint}.
Robot competitions have also produced system-level reports~\cite{krotkov2017drc,atkeson2018drc,tranzatto2022cerberus,wang2025soccer}, and a magazine report describes the hardware of a half-marathon-winning humanoid but not its autonomy~\cite{de2026marathon}.
Reports of fully autonomous humanoid track racing under real-world perception and thermal constraints remain limited.

Without developing a race-specific body, we entered a commercially
available Unitree G1 in the 100\,m, 400\,m, and 1500\,m events at the World
Humanoid Robot Games (WHRG) 2026 in Beijing.
The games were held on 22--26 August with 666 teams across all events.
Our G1 placed 13th of 74 entrants in
the 100\,m at \SI{24.3}{\second}, was 10th of 44 in the 400\,m first round at 1:55.93
before falling in the semifinal, and placed seventh of 35 in the 1500\,m at 7:43.89.

This paper makes three practical contributions: (1) integration and competition deployment of lane perception, path tracking, and learned running control on a commercially available humanoid (\cref{sec:system,sec:lane,sec:policy}); (2) image stabilization and ground-plane lane tracking against head-camera shake, and leg odometry that bridges flight phases (\cref{sec:lane});
and (3) torque-penalized training and per-event speed caps set by thermal margin (\cref{sec:operation}).

\section{Competition Setting}
\label{sec:competition}

The WHRG 2026 running events took place on an athletics track laid over an indoor speed-skating oval in Beijing,
so its curves were tighter than those of standard athletics tracks, which constrained the lookahead and turning radius (\cref{sec:lane,sec:policy}).
We entered the 100\,m and 400\,m small-class races (height up to \SI{1.4}{\meter}) and the open 1500\,m race (\cref{tab:results}).
\looseness=-1

The rules required the robot to run autonomously, with no external control after the start,
making onboard sensing the source of all steering commands (\cref{sec:lane}).
The only operator input was the start command, issued just after the official start, so the official times include this delay.

\section{System Overview}
\label{sec:system}

We used a commercially available 29-DoF Unitree G1;
\cref{fig:system} numbers the components we added or programmed.
An added Jetson AGX Orin (marker~1), which we call PC3, processes images from a head-mounted Intel RealSense D555 (marker~2) and generates the forward-speed and yaw-rate command $(v,\omega)$ (\cref{sec:lane}); the name PC3 follows the G1's built-in PC1 and PC2.
The D555 replaces the stock camera for its wider field of view
and supplies RGB and depth images at $1280\times720$ and \SI{15}{\hertz}.
The command is transmitted over Ethernet using ROS~2~\cite{macenski2022ros2} to the built-in Jetson Orin NX (PC2, marker~3), which executes the running policy and thermal speed limiter (\cref{sec:policy,sec:operation}).
As $(v,\omega)$ is the sole interface between perception and locomotion, we tuned the running policy with manually issued commands before integrating autonomy.
We also added heat-dissipation plates to the actuators.
The modifications increased the manufacturer's stated mass of \SI{35}{\kilo\gram}~\cite{unitree_g1} by \SI{1.5}{\kilo\gram}.
\looseness=-1

\section{Lane Perception and Path Tracking}
\label{sec:lane}

We adapted autonomous-driving techniques to humanoid running, where two assumptions failed: moderate camera motion (\cref{sec:lane:perception}) and continuous ground contact for odometry (\cref{sec:lane:tracking}).
Given limited on-site tuning time, we selected RGB-based lane detection rather than developing a LiDAR-based pipeline.
The head-mounted D555 provides RGB images for lane detection, Ethernet connectivity to avoid vibration-induced USB disconnections, stereo depth for ground-plane estimation, and an IMU for image stabilization and leveling.
To rehearse the venue before arrival,
we built a replica of its track in MuJoCo~\cite{todorov2012mujoco} from the venue dimensions
and ran the same ROS~2 stack, from detection to the running policy, in closed loop on rendered head-camera images.
\looseness=-1

\subsection{Lane perception}
\label{sec:lane:perception}
We qualitatively compared GANet~\cite{wang2022ganet}, BezierLaneNet~\cite{feng2022bezierlanenet}, UFLDv2~\cite{qin2024ufldv2}, and CLRerNet~\cite{honda2024clrernet} on stadium images.
CLRerNet produced the most stable lane detections for the ground-plane tracking described below and was therefore selected.
We attribute this to its use of global image context:
track lanes are spaced more narrowly than road lanes,
and keypoint-based detectors~\cite{wang2022ganet} tended to merge them in their low-resolution heatmaps.
We initialized its ResNet-34 backbone~\cite{he2016resnet} from CULane~\cite{pan2018scnn} weights and fine-tuned it on 3000 real and 300 synthetic frames.
The real frames were recorded at 30 outdoor stadiums by a person running with an action camera, not by the robot.
The synthetic indoor-venue frames were rendered from the replica and made photorealistic with an image-generation model,
because a detector trained on real frames alone made many errors on them.
A further 500 real and 200 synthetic frames were reserved for validation.
Augmentation varied brightness, color, scale, and rotation while preserving the annotated lanes' bottom-to-top direction, which the detector assumes.
The detector runs in TensorRT FP16 at \SI{17.6}{\milli\second} per frame, supporting \SI{15}{\hertz} operation.

The first failed assumption was that image motion is moderate, which running violated in two ways.
First, the head camera's roll oscillated around its intended orientation by 2.0--\SI{3.3}{\degree} root mean square (RMS) during the races,
causing the apparent lane orientations to fluctuate.
Second, a downward-tilted camera shifted the vanishing point upward relative to its position in the training images.
To compensate for both, we stabilize and level each frame before lane detection.
Following gyroscope-based video stabilization~\cite{karpenko2011gyro}, we estimate the intended orientation using a Madgwick filter~\cite{madgwick2011imu} and a low-pass filter with a \SI{0.5}{\second} time constant.
A pure-rotation homography then removes the deviation from this intended orientation.
We apply the same smoothing to the camera pose used for ground-plane projection,
preventing lane oscillations caused by small phase mismatches between the stabilized image and the estimated pose.
Training and testing without leveling visibly degraded detection, although we did not quantify this effect.
Stabilization and roll leveling were used in all races, while camera-pitch correction for the second issue was introduced before the \SI{100}{\metre} race (\cref{sec:results}).

Tracking lanes directly in the image proved unreliable because
lane detections were occasionally missed or occluded by other robots.
In the image, camera motion shifts each lane from frame to frame, so earlier detections cannot directly fill these gaps.
We therefore project detections onto the ground plane and accumulate them in the odometry frame (\cref{sec:lane:tracking}), where lanes stay nearly fixed because robot motion is compensated, so detections from successive frames can be combined.
The ground plane is estimated robustly from depth pixels with
upward-facing surface normals.
When insufficient valid depth data are available, such as in
overexposed regions, we instead estimate the plane using the
camera IMU and known camera height.
The accumulation uses a \SI{0.20}{\meter} voxel grid.
Voxel weights decay over time without new observations.
For each lane line, the accumulated points are then fitted
with a cubic polynomial extending up to \SI{18}{\meter} ahead.
Across the three logged races, the detector identified at least four
lane lines in every frame, so no frame lacked detections entirely; the accumulation only filled in lane lines that were missed or occluded.

\subsection{Path tracking}
\label{sec:lane:tracking}
The second failed assumption was that ground contact is continuous for odometry.
The velocity estimate provided with the G1 reverses sign during part of every step,
so its integrated position advances at only about \SI{1}{\meter\per\second} on average during running.
We infer that
its lagging contact detection keeps trusting the foot that has just lifted off.
Our own leg-kinematic odometry therefore assumes a non-slipping stance foot~\cite{bloesch2012legodom,hartley2018invekf},
detects stance from foot height, and bridges flight phases with the IMU.
Over a 400\,m lap of a practice track, the odometry measured \SI{377}{\meter} (a ratio of 0.94), which we judged accurate enough for tracking the path relative to the detected lanes.
At the venue, an overall ratio that also includes path-tracking offset and heading drift was about 0.90 (Appendix).

Shortly after the start, the nearest tracked lane line to the right of the camera is selected,
and the path is that line shifted toward the lane center.
Vector pursuit~\cite{wit2004vectorpursuit}, an extension of pure pursuit~\cite{coulter1992purepursuit},
tracks the path at \SI{100}{\hertz} with a \SI{6.0}{\meter} lookahead and a \SI{16.5}{\meter} minimum turning radius.
These settings kept the yaw-rate command within the \SI{\pm 0.4}{\radian\per\second} command range of the controller,
without requiring the robot to slow in curves.

\subsection{Venue adaptation}
The venue appeared in the training set only as synthetic frames rendered from a replica built to its dimensions.
The venue was darker than the outdoor stadiums in the training set,
and reflections from the ceiling lights made the brightness uneven along the track.
We therefore recorded footage at the venue, fixed the camera exposure and gain manually,
and confirmed on the footage that the detector found the lanes correctly before using it in the races.
We did not retrain the detector on site, to avoid regressions.

\section{Running Control}
\label{sec:policy}

The running policy maps commanded speed and
yaw rate to joint position targets and feedforward torques.

\subsection{Training}
\label{sec:training}
We trained the running policy in Isaac Lab~\cite{nvidia2025isaaclab}, building on Unitree's open-source training code~\cite{unitree_rl_lab}, in three stages: PPO~\cite{schulman2017ppo}, PPO with an adversarial motion prior (AMP)~\cite{peng2021amp,escontrela2022amp}, and final PPO fine-tuning.
In simulation, the hip-pitch torque saturated when policies were trained without AMP.
The AMP stage, which uses reference motions captured from human runners and retargeted as in~\cite{peng2018deepmimic,peng2020imitating,cheng2024expressive} (Appendix), reduced this saturation.
The reward included speed tracking and a penalty on squared hip-pitch torque as a proxy for resistive heating~\cite{seok2015cheetah}.
We targeted the hip pitch because it was the hottest joint in practice runs, as it also was in the 100\,m race.
Temperature was excluded from observations and rewards because actuator thermal behavior is difficult to simulate.
The policy runs at \SI{50}{\hertz} using five steps of proprioceptive, command, and action history.
It outputs position targets for 29 joints and feedforward torques for the hip-pitch and knee joints.
PC2 interpolates the position targets to \SI{1}{\kilo\hertz} for MIT control~\cite{wensing2017proprioceptive}.
We trained two policies, endurance and sprint, with forward-speed commands from 0 up to the curriculum levels reached by the deployed checkpoints, 5.46 and \SI{6.20}{\meter\per\second}, respectively (\cref{tab:results}); the endurance policy's curve commands used $\omega = v/R$ for a sampled turning radius $R$, whereas the 100\,m sprint policy, run on a straight, used none in its final training stage.
Additional training settings are provided in the Appendix.

\subsection{Hardware-aware policy design}
\begin{itemize}
  \item Low stiffness with feedforward torque on the propulsive joints.
        Prior work reports that the gains and action space shape what a policy
        learns~\cite{peng2017actionspace,chen2023learningtorque,bronars2026tunetolearn}.
        These studies examine the effects of action spaces and controller gains, but do not evaluate the combination of position targets and feedforward torque used here.
        With position targets alone, the PD stiffness must generate the propulsive torque as well as track the targets, so we expected feedforward torque to drive these joints more efficiently.
        Across the two policies, the hip-pitch and knee stiffness values were 10.0--10.7 and
        15.0--\SI{15.3}{\newton\meter\per\radian}, respectively, compared with
        \SI{100}{\newton\meter\per\radian} at hip roll and yaw.
  \item Thermal degradation in simulation.
        To account for a possible reduction of the torque constant with heating,
        we added to the dynamics randomization~\cite{peng2018dr} a per-joint first-order thermal model
        that lowers the torque constant within an episode (Appendix).
        The policy cannot observe this loss directly and must continue running despite it;
        this randomization targets robustness, not heat reduction.
\end{itemize}

\section{Thermal Management and Pacing}
\label{sec:operation}

\subsection{Heat as the key speed constraint}
Sustained speed was limited primarily by motor heating. Before mitigation, the firmware's over-temperature protection stopped the motors after approximately 50\,m of running, causing a fall.
We therefore treated thermal margin as a primary constraint on race pacing.
A half-marathon humanoid faced a similar limitation but addressed it through running-specific gearing and liquid cooling~\cite{de2026marathon}.

\subsection{Three mitigations}
\begin{itemize}
  \item \emph{Policy-level.}
        The reward penalizes torque as a proxy for heat (\cref{sec:training}).
  \item \emph{Runtime limiter.}
        We monitor the maximum motor-driver MOSFET temperature across all joints (the \emph{motor-driver temperature}).
        Above \SI{120}{\celsius}, the thermal cap decreases toward \SI{2.0}{\meter\per\second} over \SI{4}{\second}; at \SI{127}{\celsius}, which is below the observed \SI{130}{\celsius} firmware cutoff, the robot is commanded to stop.
        A race-director node on PC2 sends the \emph{speed cap}, the lower of the thermal and event-specific caps, to PC3, where path tracking uses it as the target forward speed $v$ and recomputes $\omega = v/R$ to keep the path curvature; PC2 also clips the policy's forward-speed command.
  \item \emph{Hardware.}
        Heat-dissipation plates on the hip-pitch, knee, and ankle actuators conduct motor heat to the housing (\cref{sec:system}).
\end{itemize}

\subsection{Per-event operating points}
Because heat accumulates over a race, we deliberately operated the same robot at different points on the
speed-versus-completion-margin curve:
the 100\,m spent its thermal margin on speed,
the 400\,m raised the event-specific cap only for the last straight,
and the 1500\,m held one constant cap.
Apart from the camera-pitch correction added before the 100\,m (\cref{sec:results}), two settings changed between events (\cref{tab:results}):
the policy and the event-specific cap, which was scheduled by distance, time, and detected curve state.

\section{Results at WHRG 2026}
\label{sec:results}

\begin{figure}[t]
  \centering
  \includegraphics[width=\columnwidth]{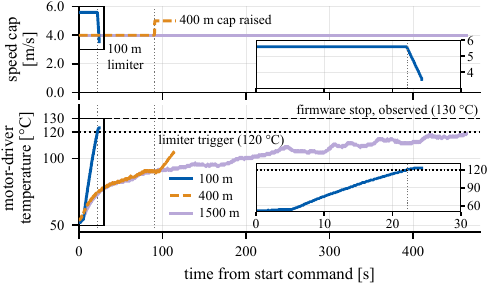}
  \vspace{-22pt}
  \caption{Speed cap (top; the lower of the event-specific and thermal caps) and motor-driver temperature (bottom; the input of the thermal limiter, \cref{sec:operation}) in the three races, from the race logs.
  Time is from the start command, which the operator issued just after the official start, and each trace spans the official race time.
  Dotted vertical lines mark the limiter engaging in the 100\,m and the 400\,m cap rising.
  Insets enlarge the 100\,m over 0--\SI{30}{\second}.}
  \label{fig:thermal}
  \vspace{-18pt}
\end{figure}

The robot finished every race it started except the 400\,m semifinal, in
which it fell (\cref{tab:results}); the motor-driver temperature remained a limiting factor (\cref{fig:thermal}).
The 400\,m and 1500\,m races were held on 23 August and the 100\,m on 25 August.

\begin{table}[t]
  \centering
  \caption[Per-event settings and results.]{\baselineskip=9pt Per-event settings and results.\par}
  \label{tab:results}
  \vspace{-6pt}
  \setlength{\tabcolsep}{3pt}
  \begin{tabular}{lccc}
    \toprule
    \noalign{\vskip -2pt}
     & 100\,m & 400\,m & 1500\,m \\[-2pt]
    \midrule
    Class                              & small & small & open \\
    Entrants                           & 74 & 44 & 35 \\
    Event-specific cap (\si{\meter\per\second}) & 5.6 & 4.0\,$\rightarrow$\,5.0 & 4.0 \\
    Policy                         & sprint & endurance & endurance \\
    Mean leg-odometry speed$^{\ast}$ (\si{\meter\per\second}) & 4.26 & 3.19\,$\rightarrow$\,3.78 & 2.98 \\
    Peak motor-driver temp.\ (\si{\celsius})$^{\dagger}$ & 123 & 106 & 119 \\
    Thermal limiter engaged & yes & no & no \\
    \midrule
    Race time (s or min:s) & \begin{tabular}[t]{@{}c@{}}24.3\\(first round)\end{tabular} & \begin{tabular}[t]{@{}c@{}}1:55.93\\(first round)\end{tabular} & 7:43.89 \\
    Final place                        & 13th & \begin{tabular}[t]{@{}c@{}}DNF\\(semifinal)\end{tabular} & 7th \\
    \bottomrule
  \end{tabular}
  \par\vspace{0.4ex}
  \parbox{\columnwidth}{\footnotesize
  $^{\ast}$Low-passed at \SI{0.5}{\hertz} and averaged over
  5--\SI{22.1}{\second} after the start command (100\,m, until the limiter engaged), 10--90 and 90--\SI{115.93}{\second} (400\,m, before and after the cap was raised),
  and 10--\SI{460}{\second} (1500\,m); not ground truth, as the scale error (\cref{sec:lane}) is not corrected.
  $^{\dagger}$Over the intervals plotted in \cref{fig:thermal}.}
  \vspace{-2ex}
\end{table}

\subsection{400\,m}
The robot qualified in 10th place in the first round and did not finish the semifinal.
In the first round, \SI{90.4}{\second} after the start command,
the event-specific cap rose from 4.0 to \SI{5.0}{\meter\per\second} for the final sprint on the last straight.
At system start-up before the semifinal, lane detection appeared abnormal, and there was no time to correct it during setup;
in the race, the robot moved diagonally to the right across its lane and fell approximately
\SI{5}{\second} after the start.
No log was recorded for this run because of a technical problem, so the cause could not be identified.
The semifinal was run in lane~1, where detection and running had worked in practice; one unverified hypothesis is that the camera mount was bent in a fall during morning practice,
tilting the camera downward and shifting the vanishing point relative to the training images.
For the 100\,m race, held on a later day, we warped each image by the camera pitch estimated from the camera IMU,
keeping the vanishing point near the center.

\subsection{1500\,m}
The motor-driver temperature rose from \SI{53}{\celsius} at the start
command and peaked \SI{460.6}{\second} later, near the end of the plotted interval, at \SI{119}{\celsius}:
\SI{1}{\celsius} below the limiter trigger and \SI{11}{\celsius} below the
observed \SI{130}{\celsius} firmware cutoff.
From 10 to \SI{460}{\second} after the start command, the logged speed command stayed at \SI{4.0}{\meter\per\second},
whereas the leg-odometry speed averaged \SI{2.98}{\meter\per\second}, \SI{1.02}{\meter\per\second} lower.
The official time gives a mean speed of \SI{3.23}{\meter\per\second} over the race distance, including the start, so the robot ran below the command; the lower odometry speed is consistent with the odometry scale error at the venue (\cref{sec:lane:tracking}).
Seven of the 35 entrants finished the race.

\subsection{100\,m}
The robot did not advance beyond the first round. Its high-speed setting reached the limiter trigger.
The motor-driver temperature rose from
\SI{52}{\celsius} at the start command and crossed the \SI{120}{\celsius}
limiter trigger \SI{22.1}{\second} after the command; the limiter began lowering the speed cap from
\SI{5.6}{\meter\per\second}, and the temperature peaked at
\SI{123}{\celsius}, \SI{4}{\celsius} below the \SI{127}{\celsius} stop command.
Because the logs are not aligned with the official timing, the time lost to the slowdown is unknown.

\section{Conclusion}
\label{sec:conclusion}

At WHRG 2026, a commercially available Unitree G1 completed three of four autonomous races; it fell in the 400\,m semifinal after a lane-detection error whose cause could not be identified.
This experience provides four practical lessons for real-world deployment.
First, perception must withstand body motion and occlusion: head-camera roll of 2.0--\SI{3.3}{\degree} RMS motivated image stabilization and ground-plane lane tracking, and lanes were detected in every logged race frame (\cref{sec:lane}).
Second, state estimation must bridge the flight phases of running, which leg odometry with IMU bridging provided (\cref{sec:lane}).
Third, adapting perception to a new venue need not require retraining on site: in the three logged races, fixing camera exposure and gain sufficed for a detector already trained on synthetic frames of the venue (\cref{sec:lane}).
Finally, reliable running benefits from hardware-aware policy training and thermal management (\cref{sec:policy,sec:operation}).
Together, these findings identify perception robustness, state estimation, venue adaptation, and thermal management as key barriers to reliable humanoid running in real-world environments.

\section*{Acknowledgment}

We thank the team members who built, tuned, and operated the robot during the competition,
the runners whose motion was captured for the reference gait,
and the organizers and venue staff of WHRG 2026.

We used Claude Code (Anthropic) to assist with the implementation of training,
motion-retargeting, and onboard software, the analysis of competition logs,
and manuscript preparation, including text and figures.
We also used Gemini (\texttt{gemini-3.1-flash-image}, Google) to make synthetic lane-detector training images photorealistic,
as described in \cref{sec:lane} and the Appendix.
We reviewed all AI-assisted content.

\vspace{-1.5ex}
\bibliographystyle{IEEEtran}
\bibliography{references}

@IEEEtranBSTCTL{IEEEexample:BSTcontrol,
  CTLuse_forced_etal       = {yes},
  CTLmax_names_forced_etal = {6},
  CTLnames_show_etal       = {1}
}

@inproceedings{bronars2026tunetolearn,
  author    = {Bronars, Antonia and Park, Younghyo and Agrawal, Pulkit},
  title     = {Tune to learn: How controller gains shape robot policy learning},
  booktitle = {Proc. Robot.: Sci. Syst. (RSS)},
  year      = {2026},
  doi       = {10.15607/RSS.2026.XXII.139}
}

@inproceedings{peng2017actionspace,
  author    = {Peng, Xue Bin and van de Panne, Michiel},
  title     = {Learning locomotion skills using {DeepRL}: Does the choice of action space matter?},
  booktitle = {Proc. ACM SIGGRAPH/Eurographics Symp. Comput. Animation (SCA)},
  pages     = {12:1--12:13},
  year      = {2017},
  doi       = {10.1145/3099564.3099567}
}

@inproceedings{chen2023learningtorque,
  author    = {Chen, Shuxiao and Zhang, Bike and Mueller, Mark W. and Rai, Akshara and Sreenath, Koushil},
  title     = {Learning torque control for quadrupedal locomotion},
  booktitle = {Proc. IEEE-RAS Int. Conf. Humanoid Robots (Humanoids)},
  pages     = {1--8},
  year      = {2023},
  doi       = {10.1109/Humanoids57100.2023.10375154}
}

@misc{wei2026sprint,
  author = {Wei, Yantong and Huang, Kaihong and Pan, Hainan and Luo, Jiawei and Zhou, Jiawei and Mai, Ziyan and Zeng, Zhiwen and Wang, Yaonan and Lu, Huimin},
  title  = {{SPRINT}: Efficient spectral priors for humanoid athletic sprints},
  year   = {2026},
  note   = {arXiv:2605.28549},
  doi    = {10.48550/arXiv.2605.28549}
}

@misc{de2026marathon,
  author       = {De, Avik},
  title        = {The secret to marathon-winning humanoid robots},
  howpublished = {IEEE Spectrum},
  month        = {Jun. 17,},
  year         = {2026},
  note         = {{A}ccessed: Jun. 26, 2026},
  url          = {https://spectrum.ieee.org/china-humanoid-robot-marathon}
}

@inproceedings{rudin2022minutes,
  author    = {Rudin, Nikita and Hoeller, David and Reist, Philipp and Hutter, Marco},
  title     = {Learning to walk in minutes using massively parallel deep reinforcement learning},
  booktitle = {Proc. Conf. Robot Learn. (CoRL)},
  series    = {PMLR},
  volume    = {164},
  pages     = {91--100},
  year      = {2022}
}

@misc{schulman2017ppo,
  author = {Schulman, John and Wolski, Filip and Dhariwal, Prafulla and Radford, Alec and Klimov, Oleg},
  title  = {Proximal policy optimization algorithms},
  year   = {2017},
  note   = {arXiv:1707.06347},
  doi    = {10.48550/arXiv.1707.06347}
}

@misc{unitree_rl_lab,
  author       = {{Unitree Robotics}},
  title        = {{Unitree RL Lab}},
  howpublished = {GitHub},
  year         = {2025},
  note         = {{A}ccessed: Jun. 12, 2026},
  url          = {https://github.com/unitreerobotics/unitree_rl_lab}
}

@inproceedings{honda2024clrernet,
  author    = {Honda, Hiroto and Uchida, Yusuke},
  title     = {{CLRerNet}: Improving confidence of lane detection with {LaneIoU}},
  booktitle = {Proc. IEEE/CVF Winter Conf. Appl. Comput. Vis. (WACV)},
  pages     = {1165--1174},
  year      = {2024},
  doi       = {10.1109/WACV57701.2024.00121}
}

@inproceedings{madgwick2011imu,
  author    = {Madgwick, Sebastian O. H. and Harrison, Andrew J. L. and Vaidyanathan, Ravi},
  title     = {Estimation of {IMU} and {MARG} orientation using a gradient descent algorithm},
  booktitle = {Proc. IEEE Int. Conf. Rehabil. Robot. (ICORR)},
  pages     = {1--7},
  year      = {2011},
  doi       = {10.1109/ICORR.2011.5975346}
}

@article{wit2004vectorpursuit,
  author    = {Wit, Jeff and Crane, Carl D. and Armstrong, David},
  title     = {Autonomous ground vehicle path tracking},
  journal   = {J. Robot. Syst.},
  volume    = {21},
  number    = {8},
  pages     = {439--449},
  month   = aug,
  year      = {2004},
  doi       = {10.1002/rob.20031}
}

@article{peng2021amp,
  author  = {Peng, Xue Bin and Ma, Ze and Abbeel, Pieter and Levine, Sergey and Kanazawa, Angjoo},
  title   = {{AMP}: Adversarial motion priors for stylized physics-based character control},
  journal = {ACM Trans. Graph.},
  volume  = {40},
  number  = {4},
  pages   = {144:1--144:20},
  month   = aug,
  year    = {2021},
  doi     = {10.1145/3450626.3459670}
}

@misc{unitree_g1,
  author       = {{Unitree Robotics}},
  title        = {{Unitree G1} humanoid robot},
  note         = {{A}ccessed: Sep. 18, 2026},
  url          = {https://www.unitree.com/g1/}
}

@article{hwangbo2019anymal,
  author  = {Hwangbo, Jemin and Lee, Joonho and Dosovitskiy, Alexey and Bellicoso, Dario and Tsounis, Vassilios and Koltun, Vladlen and Hutter, Marco},
  title   = {Learning agile and dynamic motor skills for legged robots},
  journal = {Sci. Robot.},
  volume  = {4},
  number  = {26},
  pages   = {eaau5872},
  month   = jan,
  year    = {2019},
  doi     = {10.1126/scirobotics.aau5872}
}

@article{radosavovic2024humanoid,
  author  = {Radosavovic, Ilija and Xiao, Tete and Zhang, Bike and Darrell, Trevor and Malik, Jitendra and Sreenath, Koushil},
  title   = {Real-world humanoid locomotion with reinforcement learning},
  journal = {Sci. Robot.},
  volume  = {9},
  number  = {89},
  pages   = {eadi9579},
  month   = apr,
  year    = {2024},
  doi     = {10.1126/scirobotics.adi9579}
}

@article{li2025cassie,
  author  = {Li, Zhongyu and Peng, Xue Bin and Abbeel, Pieter and Levine, Sergey and Berseth, Glen and Sreenath, Koushil},
  title   = {Reinforcement learning for versatile, dynamic, and robust bipedal locomotion control},
  journal = {Int. J. Robot. Res.},
  volume  = {44},
  number  = {5},
  pages   = {840--888},
  month   = apr,
  year    = {2025},
  doi     = {10.1177/02783649241285161}
}

@inproceedings{pan2018scnn,
  author    = {Pan, Xingang and Shi, Jianping and Luo, Ping and Wang, Xiaogang and Tang, Xiaoou},
  title     = {Spatial as deep: Spatial {CNN} for traffic scene understanding},
  booktitle = {Proc. AAAI Conf. Artif. Intell.},
  volume    = {32},
  number    = {1},
  pages     = {7276--7283},
  year      = {2018},
  doi       = {10.1609/aaai.v32i1.12301}
}

@inproceedings{zheng2022clrnet,
  author    = {Zheng, Tu and Huang, Yifei and Liu, Yang and Tang, Wenjian and Yang, Zheng and Cai, Deng and He, Xiaofei},
  title     = {{CLRNet}: Cross layer refinement network for lane detection},
  booktitle = {Proc. IEEE/CVF Conf. Comput. Vis. Pattern Recognit. (CVPR)},
  pages     = {888--897},
  year      = {2022},
  doi       = {10.1109/CVPR52688.2022.00097}
}

@incollection{atkeson2018drc,
  author    = {Atkeson, Christopher G. and Benzun, P. W. Babu and Banerjee, Nandan and Berenson, Dmitry and Bove, Christoper P. and Cui, Xiongyi and DeDonato, Mathew and Du, Ruixiang and Feng, Siyuan and Franklin, Perry and Gennert, M. and Graff, Joshua P. and He, Peng and Jaeger, Aaron and Kim, Joohyung and Knoedler, Kevin and Li, Lening and Liu, Chenggang and Long, Xianchao and Padir, T. and Polido, Felipe and Tighe, G. G. and Xinjilefu, X.},
  title     = {What happened at the {DARPA Robotics Challenge} finals},
  booktitle = {The {DARPA} Robotics Challenge Finals: Humanoid Robots To The Rescue},
  editor    = {Spenko, Matthew and Buerger, Stephen and Iagnemma, Karl},
  series    = {Springer Tracts in Advanced Robotics},
  volume    = {121},
  publisher = {Springer},
  address   = {Cham, Switzerland},
  pages     = {667--684},
  year      = {2018},
  doi       = {10.1007/978-3-319-74666-1_17}
}

@misc{nvidia2025isaaclab,
  author = {Mittal, Mayank and Roth, Pascal and Tigue, James and Richard, Antoine and Zhang, Octi and Du, Peter and Serrano-Mu{\~n}oz, Antonio and Yao, Xinjie and Zurbr{\"u}gg, Ren{\'e} and Rudin, Nikita and Wawrzyniak, Lukasz and Rakhsha, Milad and Denzler, Alain and Heiden, Eric and Borovicka, Ales and Ahmed, Ossama and Akinola, Iretiayo and Anwar, Abrar and Carlson, Mark T. and Feng, Ji Yuan and Garg, Animesh and Gasoto, Renato and Gulich, Lionel and Guo, Yijie and Gussert, M. and Hansen, Alex and Kulkarni, Mihir and Li, Chenran and Liu, Wei and Makoviychuk, Viktor and Malczyk, Grzegorz and Mazhar, Hammad and Moghani, Masoud and Murali, Adithyavairavan and Noseworthy, Michael and Poddubny, Alexander and Ratliff, Nathan and Rehberg, Welf and Schwarke, Clemens and Singh, Ritvik and Smith, James Latham and Tang, Bingjie and Thaker, Ruchik and Trepte, Matthew and Van Wyk, Karl and Yu, Fangzhou and Millane, Alex and Ramasamy, Vikram and Steiner, Remo and Subramanian, Sangeeta and Volk, Clemens and Chen, CY and Jawale, Neel and Kuruttukulam, Ashwin Varghese and Lin, Michael A. and Mandlekar, Ajay and Patzwaldt, Karsten and Welsh, John and Zhao, Huihua and Anes, Fatima and Lafleche, Jean-Francois and Mo{\"e}nne-Loccoz, Nicolas and Park, Soowan and Stepinski, Rob and Van Gelder, Dirk and Amevor, Chris and Carius, Jan and Chang, Jumyung and Chen, Anka He and de Heras Ciechomski, Pablo and Daviet, Gilles and Mohajerani, Mohammad and von Muralt, Julia and Reutskyy, Viktor and Sauter, Michael and Schirm, Simon and Shi, Eric L. and Terdiman, Pierre and Vilella, Kenny and Widmer, Tobias and Yeoman, Gordon and Chen, Tiffany and Grizan, Sergey and Li, Cathy and Li, Lotus and Smith, Connor and Wiltz, Rafael and Alexis, Kostas and Chang, Yan and Chu, David and Fan, Linxi and Farshidian, Farbod and Handa, Ankur and Huang, Spencer and Hutter, Marco and Narang, Yashraj and Pouya, Soha and Sheng, Shiwei and Zhu, Yuke and Macklin, Miles and Moravanszky, Adam and Reist, Philipp and Guo, Yunrong and Hoeller, David and State, Gavriel},
  title  = {{Isaac Lab}: A {GPU}-accelerated simulation framework for multi-modal robot learning},
  year   = {2025},
  note   = {arXiv:2511.04831},
  doi    = {10.48550/arXiv.2511.04831}
}

@inproceedings{escontrela2022amp,
  author    = {Escontrela, Alejandro and Peng, Xue Bin and Yu, Wenhao and Zhang, Tingnan and Iscen, Atil and Goldberg, Ken and Abbeel, Pieter},
  title     = {Adversarial motion priors make good substitutes for complex reward functions},
  booktitle = {Proc. IEEE/RSJ Int. Conf. Intell. Robots Syst. (IROS)},
  pages     = {25--32},
  year      = {2022},
  doi       = {10.1109/IROS47612.2022.9981973}
}

@article{seok2015cheetah,
  author  = {Seok, Sangok and Wang, Albert and Chuah, Meng Yee Michael and Hyun, Dong Jin and Lee, Jongwoo and Otten, David M. and Lang, Jeffrey H. and Kim, Sangbae},
  title   = {Design principles for energy-efficient legged locomotion and implementation on the {MIT} {Cheetah} robot},
  journal = {IEEE/ASME Trans. Mechatronics},
  volume  = {20},
  number  = {3},
  pages   = {1117--1129},
  month   = jun,
  year    = {2015},
  doi     = {10.1109/TMECH.2014.2339013}
}

@techreport{karpenko2011gyro,
  author      = {Karpenko, Alexandre and Jacobs, David and Baek, Jongmin and Levoy, Marc},
  title       = {Digital video stabilization and rolling shutter correction using gyroscopes},
  institution = {Stanford Univ.},
  type        = {Comput. Sci. Tech. Rep.},
  number      = {CSTR 2011-03},
  year        = {2011}
}

@inproceedings{siekmann2021gaits,
  author    = {Siekmann, Jonah and Godse, Yesh and Fern, Alan and Hurst, Jonathan},
  title     = {Sim-to-real learning of all common bipedal gaits via periodic reward composition},
  booktitle = {Proc. IEEE Int. Conf. Robot. Autom. (ICRA)},
  pages     = {7309--7315},
  year      = {2021},
  doi       = {10.1109/ICRA48506.2021.9561814}
}

@inproceedings{he2025asap,
  author    = {He, Tairan and Gao, Jiawei and Xiao, Wenli and Zhang, Yuanhang and Wang, Zi and Wang, Jiashun and Luo, Zhengyi and He, Guanqi and Sobanbabu, Nikhil and Pan, Chaoyi and Yi, Zeji and Qu, Guannan and Kitani, Kris and Hodgins, Jessica K. and Fan, Linxi and Zhu, Yuke and Liu, Changliu and Shi, Guanya},
  title     = {{ASAP}: Aligning simulation and real-world physics for learning agile humanoid whole-body skills},
  booktitle = {Proc. Robot.: Sci. Syst. (RSS)},
  year      = {2025},
  doi       = {10.15607/RSS.2025.XXI.066}
}

@article{krotkov2017drc,
  author  = {Krotkov, Eric and Hackett, Douglas and Jackel, Larry and Perschbacher, Michael and Pippine, James and Strauss, Jesse and Pratt, Gill and Orlowski, Christopher},
  title   = {The {DARPA Robotics Challenge} finals: Results and perspectives},
  journal = {J. Field Robot.},
  volume  = {34},
  number  = {2},
  pages   = {229--240},
  month   = mar,
  year    = {2017},
  doi     = {10.1002/rob.21683}
}

@article{peng2018deepmimic,
  author  = {Peng, Xue Bin and Abbeel, Pieter and Levine, Sergey and van de Panne, Michiel},
  title   = {{DeepMimic}: Example-guided deep reinforcement learning of physics-based character skills},
  journal = {ACM Trans. Graph.},
  volume  = {37},
  number  = {4},
  pages   = {143:1--143:14},
  month   = aug,
  year    = {2018},
  doi     = {10.1145/3197517.3201311}
}

@inproceedings{peng2020imitating,
  author    = {Peng, Xue Bin and Coumans, Erwin and Zhang, Tingnan and Lee, Tsang-Wei and Tan, Jie and Levine, Sergey},
  title     = {Learning agile robotic locomotion skills by imitating animals},
  booktitle = {Proc. Robot.: Sci. Syst. (RSS)},
  year      = {2020},
  doi       = {10.15607/RSS.2020.XVI.064}
}

@inproceedings{cheng2024expressive,
  author    = {Cheng, Xuxin and Ji, Yandong and Chen, Junming and Yang, Ruihan and Yang, Ge and Wang, Xiaolong},
  title     = {Expressive whole-body control for humanoid robots},
  booktitle = {Proc. Robot.: Sci. Syst. (RSS)},
  year      = {2024},
  doi       = {10.15607/RSS.2024.XX.107}
}

@inproceedings{peng2018dr,
  author    = {Peng, Xue Bin and Andrychowicz, Marcin and Zaremba, Wojciech and Abbeel, Pieter},
  title     = {Sim-to-real transfer of robotic control with dynamics randomization},
  booktitle = {Proc. IEEE Int. Conf. Robot. Autom. (ICRA)},
  pages     = {3803--3810},
  year      = {2018},
  doi       = {10.1109/ICRA.2018.8460528}
}

@article{wensing2017proprioceptive,
  author  = {Wensing, Patrick M. and Wang, Albert and Seok, Sangok and Otten, David and Lang, Jeffrey and Kim, Sangbae},
  title   = {Proprioceptive actuator design in the {MIT} {Cheetah}: Impact mitigation and high-bandwidth physical interaction for dynamic legged robots},
  journal = {IEEE Trans. Robot.},
  volume  = {33},
  number  = {3},
  pages   = {509--522},
  month   = jun,
  year    = {2017},
  doi     = {10.1109/TRO.2016.2640183}
}

@inproceedings{bloesch2012legodom,
  author    = {Bloesch, Michael and Hutter, Marco and Hoepflinger, Mark and Leutenegger, Stefan and Gehring, Christian and Remy, C. David and Siegwart, Roland},
  title     = {State estimation for legged robots: Consistent fusion of leg kinematics and {IMU}},
  booktitle = {Proc. Robot.: Sci. Syst. (RSS)},
  year      = {2012},
  doi       = {10.15607/RSS.2012.VIII.003}
}

@inproceedings{hartley2018invekf,
  author    = {Hartley, Ross and Ghaffari Jadidi, Maani and Grizzle, Jessy and Eustice, Ryan M.},
  title     = {Contact-aided invariant extended {Kalman} filtering for legged robot state estimation},
  booktitle = {Proc. Robot.: Sci. Syst. (RSS)},
  year      = {2018},
  doi       = {10.15607/RSS.2018.XIV.050}
}

@inproceedings{he2016resnet,
  author    = {He, Kaiming and Zhang, Xiangyu and Ren, Shaoqing and Sun, Jian},
  title     = {Deep residual learning for image recognition},
  booktitle = {Proc. IEEE Conf. Comput. Vis. Pattern Recognit. (CVPR)},
  pages     = {770--778},
  year      = {2016},
  doi       = {10.1109/CVPR.2016.90}
}

@techreport{coulter1992purepursuit,
  author      = {Coulter, R. Craig},
  title       = {Implementation of the pure pursuit path tracking algorithm},
  institution = {Robotics Inst., Carnegie Mellon Univ.},
  address     = {Pittsburgh, PA, USA},
  type        = {Tech. Rep.},
  number      = {CMU-RI-TR-92-01},
  month       = jan,
  year        = {1992}
}

@article{macenski2022ros2,
  author  = {Macenski, Steven and Foote, Tully and Gerkey, Brian and Lalancette, Chris and Woodall, William},
  title   = {{Robot Operating System 2}: Design, architecture, and uses in the wild},
  journal = {Sci. Robot.},
  volume  = {7},
  number  = {66},
  pages   = {eabm6074},
  month   = may,
  year    = {2022},
  doi     = {10.1126/scirobotics.abm6074}
}

@inproceedings{todorov2012mujoco,
  author    = {Todorov, Emanuel and Erez, Tom and Tassa, Yuval},
  title     = {{MuJoCo}: A physics engine for model-based control},
  booktitle = {Proc. IEEE/RSJ Int. Conf. Intell. Robots Syst. (IROS)},
  pages     = {5026--5033},
  year      = {2012},
  doi       = {10.1109/IROS.2012.6386109}
}

@inproceedings{qin2020ufld,
  author    = {Qin, Zequn and Wang, Huanyu and Li, Xi},
  title     = {Ultra fast structure-aware deep lane detection},
  booktitle = {Proc. Eur. Conf. Comput. Vis. (ECCV)},
  pages     = {276--291},
  year      = {2020},
  doi       = {10.1007/978-3-030-58586-0_17}
}

@inproceedings{feng2022bezierlanenet,
  author    = {Feng, Zhengyang and Guo, Shaohua and Tan, Xin and Xu, Ke and Wang, Min and Ma, Lizhuang},
  title     = {Rethinking efficient lane detection via curve modeling},
  booktitle = {Proc. IEEE/CVF Conf. Comput. Vis. Pattern Recognit. (CVPR)},
  pages     = {17041--17049},
  year      = {2022},
  doi       = {10.1109/CVPR52688.2022.01655}
}

@article{qin2024ufldv2,
  author    = {Qin, Zequn and Zhang, Pengyi and Li, Xi},
  title     = {Ultra fast deep lane detection with hybrid anchor driven ordinal classification},
  journal   = {IEEE Trans. Pattern Anal. Mach. Intell.},
  volume    = {46},
  number    = {5},
  pages     = {2555--2568},
  month   = may,
  year      = {2024},
  doi       = {10.1109/TPAMI.2022.3182097}
}

@inproceedings{wang2022ganet,
  author    = {Wang, Jinsheng and Ma, Yinchao and Huang, Shaofei and Hui, Tianrui and Wang, Fei and Qian, Chen and Zhang, Tianzhu},
  title     = {A keypoint-based global association network for lane detection},
  booktitle = {Proc. IEEE/CVF Conf. Comput. Vis. Pattern Recognit. (CVPR)},
  pages     = {1382--1391},
  year      = {2022},
  doi       = {10.1109/CVPR52688.2022.00145}
}

@inproceedings{agarwal2023egocentric,
  author    = {Agarwal, Ananye and Kumar, Ashish and Malik, Jitendra and Pathak, Deepak},
  title     = {Legged locomotion in challenging terrains using egocentric vision},
  booktitle = {Proc. Conf. Robot Learn. (CoRL)},
  series    = {PMLR},
  volume    = {205},
  pages     = {403--415},
  year      = {2023}
}

@inproceedings{fu2022coupling,
  author    = {Fu, Zipeng and Kumar, Ashish and Agarwal, Ananye and Qi, Haozhi and Malik, Jitendra and Pathak, Deepak},
  title     = {Coupling vision and proprioception for navigation of legged robots},
  booktitle = {Proc. IEEE/CVF Conf. Comput. Vis. Pattern Recognit. (CVPR)},
  pages     = {17252--17262},
  year      = {2022},
  doi       = {10.1109/CVPR52688.2022.01676}
}

@article{tranzatto2022cerberus,
  author  = {Tranzatto, Marco and Mascarich, Frank and Bernreiter, Lukas and Godinho, Carolina and Camurri, Marco and Khattak, Shehryar and Dang, Tung and Reijgwart, Victor and L{\"o}je, Johannes and Wisth, David and Zimmermann, Samuel and Nguyen, Huan and Fehr, Marius and Solanka, Lukas and Buchanan, Russell and Bjelonic, Marko and Khedekar, Nikhil and Valceschini, Mathieu and Jenelten, Fabian and Dharmadhikari, Mihir and Homberger, Timon and De Petris, Paolo and Wellhausen, Lorenz and Kulkarni, Mihir and Miki, Takahiro and Hirsch, Satchel and Montenegro, Markus and Papachristos, Christos and Tresoldi, Fabian and Carius, Jan and Valsecchi, Giorgio and Lee, Joonho and Meyer, Konrad and Wu, Xiangyu and Nieto, Juan and Smith, Andy and Hutter, Marco and Siegwart, Roland and Mueller, Mark and Fallon, Maurice and Alexis, Kostas},
  title   = {{CERBERUS}: Autonomous legged and aerial robotic exploration in the tunnel and urban circuits of the {DARPA Subterranean Challenge}},
  journal = {Field Robot.},
  volume  = {2},
  number  = {1},
  pages   = {274--324},
  month   = mar,
  year    = {2022},
  doi     = {10.55417/fr.2022011}
}

@misc{wang2025soccer,
  author = {Wang, Quanyou and Zhu, Mingzhang and Hou, Ruochen and Gillespie, Kay and Zhu, Alvin and Wang, Shiqi and Wang, Yicheng and Fernandez, Gaberiel I. and Liu, Yeting and Togashi, Colin and Nam, Hyunwoo and Navghare, Aditya and Xu, Alex and Zhu, Taoyuanmin and Ahn, Min Sung and Flores Alvarez, Arturo and Quan, Justin and Hong, Ethan and Hong, Dennis W.},
  title  = {A hierarchical, model-based system for high-performance humanoid soccer},
  year   = {2025},
  note   = {arXiv:2512.09431},
  doi    = {10.48550/arXiv.2512.09431}
}

\twocolumn[{\centering\fontsize{16pt}{19pt}\selectfont Appendix: Supplementary Technical Details\par\vspace{1.5em}}]
\typeout{APPENDIX-START-PAGE \thepage}
\setcounter{table}{0}
\renewcommand{\thetable}{S\arabic{table}}
This appendix gives technical details that supplement the paper.

\section*{Leg Odometry Implementation and Accuracy}
Our odometry detects stance when a foot is within \SI{2}{\centi\meter} of its lowest height over the preceding \SI{0.8}{\second}.
During stance, foot-kinematic velocity and IMU-based prediction are combined with weights of 0.8 and 0.2, respectively; during flight, the estimate relies on the IMU alone.
On a standard \SI{400}{\meter} practice track, the odometry measured \SI{377}{\meter} per lap, corresponding to a distance scale of 0.94.
At the competition venue, we further assessed its consistency using the recorded trajectories.
Because heading drift rotated the estimated oval between laps, we segmented laps by accumulated heading change and fitted circles to individual curve passes.
In the 1500\,m race, three complete laps measured 352.3, 345.1, and \SI{349.3}{\meter}, with a standard deviation of \SI{1.0}{\percent}.
The estimated curve radius averaged \SI{24.20}{\meter} across seven passes.
The corresponding lane-center dimensions were \SI{387.3}{\meter} per lap and \SI{26.49}{\meter} in radius, giving an overall scale ratio of approximately 0.90.
This ratio includes not only odometry error but also path-tracking offset, heading drift, and circle-fitting approximation.

\section*{Lane Detector Training and Tracking}
CLRerNet uses a ResNet-34 backbone initialized from CULane weights, with an input resolution of $960\times960$.
The training and validation sets contained 3000 and 500 real stadium frames, respectively, and 300 and 200 synthetic frames.
The synthetic frames reproduced indoor-venue features absent from the outdoor training images, including ceiling lights, advertising boards, and spectators.
They were rendered using a replica of the competition track and made photorealistic with Gemini (\texttt{gemini-3.1-flash-image}, Google).
We did not quantify the contribution of synthetic images to detection performance at the venue.
For ground-plane lane tracking, voxel weights decay by a factor of 0.95 per processed frame, and each lane line is fitted once at least eight voxels are available.
Voxels are keyed by their position in the odometry frame and are not moved afterward; voxels more than \SI{5}{\meter} behind the camera, \SI{18}{\meter} ahead of it, or \SI{8}{\meter} to either side are discarded.
If the camera position jumps by \SI{2}{\meter} or more between frames, which we treat as an odometry reset, all accumulated voxels are discarded.
Lane IDs are fixed approximately \SI{2}{\second} after the start.
When no lanes are detected, the last estimated path is retained.

\section*{Running Policy Training}
\begin{table}[!t]
\centering\footnotesize
\caption{Final-stage PPO settings}
\label{tab:t1_ppo}
\begin{tabular}{@{}ll@{}}
\toprule
Simulator / RL library & Isaac Lab / RSL-RL \\
Physics step / policy rate & \SI{5}{\milli\second} / \SI{50}{\hertz} \\
Episode length & \SI{40}{\second} \\
Actor and critic & 512--256--128, ELU \\
Parallel environments & 4096 (each policy) \\
Learning rate & \num{3e-4}, adaptive (KL target 0.01) \\
Discount factor $\gamma$ / GAE $\lambda$ & 0.99 / 0.95 \\
PPO clipping ratio & 0.2 \\
Epochs / mini-batches per update & 5 / 4 \\
Rollout steps per environment per update & 24 \\
Entropy-bonus coefficient & 0.01 \\
\bottomrule
\end{tabular}
\end{table}

\begin{table}[!t]
\centering\footnotesize
\caption{Final-stage settings that differ between the policies}
\label{tab:t1_policies}
\begin{tabular}{@{}>{\raggedright\arraybackslash}p{\dimexpr0.30\columnwidth-\tabcolsep\relax}>{\raggedright\arraybackslash}p{\dimexpr0.33\columnwidth-\tabcolsep\relax}>{\raggedright\arraybackslash}p{\dimexpr0.37\columnwidth-2\tabcolsep\relax}@{}}
\toprule
& Sprint & Endurance \\
\midrule
Forward-speed command, curriculum level reached & \SI{6.20}{\meter\per\second} & \SI{5.46}{\meter\per\second} \\
\addlinespace[2pt]
Curve commands & None & A quarter of commands; radius 18--\SI{80}{\meter}, either direction \\
\addlinespace[2pt]
Curve-speed command, curriculum level reached & -- & \SI{3.90}{\meter\per\second} \\
\addlinespace[2pt]
Squared hip-pitch torque weight & \num{-1e-4}, reduced linearly to 0.3 times between 5.0 and \SI{6.5}{\meter\per\second} & \num{-5e-4}, raised linearly from 1$\times$ at \SI{2.5}{\meter\per\second} to 6$\times$ at \SI{0}{\meter\per\second} \\
\addlinespace[2pt]
Squared knee / hip-yaw / ankle torque weights & \num{-1e-4} / \num{-1e-4} / \num{-2e-4} & \num{-4e-4} / \num{-2e-4} / \num{-2e-4} \\
\addlinespace[2pt]
Hip-pitch / knee stiffness (\si{\newton\meter\per\radian}) & 10.7 / 15.0 & 10.0 / 15.3 \\
\addlinespace[2pt]
Hip-pitch / knee damping (\si{\newton\meter\second\per\radian}) & 2.69 / 5.49 & 3.0 / 5.56 \\
\bottomrule
\end{tabular}
\end{table}

\begin{table}[!t]
\centering\footnotesize
\caption{Dynamics randomization and thermal model}
\label{tab:t1_dr}
\begin{tabular}{@{}>{\raggedright\arraybackslash}p{0.48\columnwidth}>{\raggedright\arraybackslash}p{\dimexpr0.52\columnwidth-2\tabcolsep\relax}@{}}
\toprule
Friction & 0.6--1.1 \\
Mass, joint friction, armature & $\times$0.8--1.2 \\
Torque constant & $\times$0.9--1.1 at each reset \\
\midrule
Heating / cooling time constant & 5 / \SI{120}{\second} \\
Torque-constant drop at full heating & \SI{28}{\percent} $\times$ per-joint factor 0.6--1.4 (16.8--\SI{39.2}{\percent}) \\
Heating susceptibility & $\times$0.6--1.4 per joint \\
\bottomrule
\end{tabular}
\end{table}

\subsection*{Reference motion}
For the AMP stage (\cref{sec:training}), we captured seven 50\,m sprints by two human runners at \SI{240}{\hertz} using an IMU-based motion-capture suit.
Following the motion-imitation methods cited in \cref{sec:training}, joint angles were retargeted relative to a reference posture, and root motion was scaled by the leg-length ratio (0.85--0.91) to avoid crouching.

\subsection*{Training settings}
The following settings apply to the final PPO stage (\cref{tab:t1_ppo}).
The sprint and endurance policies differed in their motion-prior stages and in the final-stage settings of \cref{tab:t1_policies}.
Curricula raised the forward-speed and curve-speed commands; \cref{tab:t1_policies} lists the levels reached by the deployed checkpoints.

\subsection*{Rewards and commands}
In the final stage, only the sprint policy, used on the straight 100\,m, received no curve commands (\cref{tab:t1_policies}).
The reward penalized squared hip-pitch torque as a proxy for heating, without using temperature.
Squared knee, hip-yaw, and ankle torques were also penalized, so that the load did not simply shift from the hip pitch to other joints.
\Cref{tab:t1_policies} gives these torque weights for each policy.
An energy penalty based on $\sum_j |\dot q_j||\tau_j|$ across all 29 joints varied from \num{-2e-5} for walking to \num{-5e-6} for running, through a sigmoid of the commanded speed that switches between 1 and \SI{5}{\meter\per\second} (midpoint \SI{3}{\meter\per\second}, slope \SI{2}{\second\per\meter}).
The L1 feedforward-torque penalty had a weight of \num{-1e-4}.

\subsection*{Dynamics randomization}
\Cref{tab:t1_dr} lists the randomized parameters and the constants of the thermal model.
The thermal model described in \cref{sec:policy} reduced the torque constant according to a first-order model driven by the squared ratio of joint torque to its limit.
The per-joint factors in \cref{tab:t1_dr} were drawn uniformly.
The model scaled each joint's total output torque.
The \SI{28}{\percent} reduction and both time constants were assumed rather than measured.
Episodes terminated when the base height fell below \SI{0.2}{\meter} or the tilt exceeded \SI{0.8}{\radian}.

\section*{Speed Limits at Deployment}
The thermal cap (\cref{sec:operation}) recovers over \SI{3}{\second} once the motor-driver temperature falls back to \SI{90}{\celsius}, which never occurred during a race.
Path tracking changes the target forward speed $v$ at no more than \SI{2.0}{\meter\per\second\squared} upward and \SI{1.0}{\meter\per\second\squared} downward.

\end{document}